\documentclass[10pt]{article}
\usepackage[table]{xcolor}
\usepackage{url}
\usepackage{naacl-hlt}
\usepackage{times}
\usepackage{latexsym}
\usepackage{amsmath,epsfig}
\usepackage{longtable}
\usepackage{makeidx}
\usepackage{graphicx}
\usepackage[american]{babel}
\usepackage{thumbpdf}
\usepackage{fancyhdr}
\usepackage{rotating}
\usepackage{color}
\usepackage{fp}
\usepackage{amssymb}
\usepackage{subfigure}
\usepackage{float}
\usepackage{verbatim}
\usepackage{multirow}
\usepackage{longtable}
\usepackage{amssymb, amsthm, mathrsfs,yhmath, slashed, wasysym, algorithm, algpseudocode, listings}

\newcommand{ \rn }[1]{\mathbb{R}^{#1}}

\title{Statistically adaptive learning for a general class of \\ cost functions (SA L-BFGS)\footnotemark[1]}

\author{\small{Stephen Purpura \footnotemark[2]\, \footnotemark[4]} \\ \small{spurpura@contextrelevant.com} \\  \small{\bf Jim Walsh \footnotemark[4]} \\ \small{jwalsh@contextrelevant.com}\And \small{\bf Dustin Hillard \footnotemark[4] } \\ \small{dhillard@contextrelevant.com} \\ \small{\bf Scott Golder \footnotemark[4]} \\ \small{sgolder@contextrelevant.com} \And \small{\bf Mark Hubenthal \footnotemark[3]\, \footnotemark[4]} \\ \small{mhubenthal@contextrelevant.com}\\  \small{\bf Scott Smith \footnotemark[4]} \\ \small{ssmith@contextrelevant.com} }

\date{}

\begin{document}
\maketitle
\renewcommand*{\thefootnote}{\fnsymbol{footnote}}

\footnotetext[1]{The material of this work is patent pending.}
\footnotetext[2]{in absentia at Department of Information Science, Cornell University, Ithaca, NY, 14850}
\footnotetext[3]{recent graduate of Department of Mathematics, University of Washington, Seattle, WA, 98195}
\footnotetext[4]{Context Relevant, Inc., Seattle, WA, 98121}

\renewcommand*{\thefootnote}{\arabic{footnote}}

\begin{abstract}
We present a system that enables rapid model experimentation for tera-scale machine learning with trillions of non-zero features, billions of training examples, and millions of parameters. Our contribution to the literature is a new method (SA L-BFGS) for changing batch L-BFGS to perform in near real-time by using statistical tools to balance the contributions of previous weights, old training examples, and new training examples to achieve fast convergence with few iterations. The result is, to our knowledge, the most scalable and flexible linear learning system reported in the literature, beating standard practice with the current best system (Vowpal Wabbit and AllReduce). Using the KDD Cup 2012 data set from Tencent, Inc. we provide experimental results to verify the performance of this method.
\end{abstract}

\section{Introduction}
\label{sec:intro}
The demand for analysis and predictive modeling derived from very large data sets has grown immensely in recent years. One of the big problems in meeting this demand is the fact that data has grown faster than the availability of raw computational speed. As such, it has been necessary to use intelligent and efficient approaches when tackling the data training process. Specifically, there has been much focus on problems of the form
\begin{equation}\label{eq:costfunction}
\min_{\theta \in \mathbb{R}^{l}}\sum_{i=1}^{m}l(\theta^{T}x^{(i)}; y^{(i)}) + \lambda S(\theta),
\end{equation}
where $x^{(i)} \in \rn{l}$ is the feature vector of the $i$th example, $y^{(i)} \in \{0,1\}$ is the label, $\theta \in \rn{l}$ is the vector of fitting parameters, $l$ is a smooth convex loss function and $S$ a regularizer. Some of the more popular methods for determining $\theta$ include linear and logistic regression, respectively. The optimal such $\theta$ corresponds to a linear predictor function $p_{\theta}(x) = \theta^{T}x$ that \textit{best} fits the data in some appropriate sense, depending on $l$ and $S$. We remark that such cost functions in (\ref{eq:costfunction}) have a structure which is naturally decomposable over the given training examples, so that all computations can potentially be run in parallel over a distributed environment. 

However, in practice it is often the case that the model must be updated accordingly as new data is acquired. That is, we want to answer the question of how $\theta$ changes in the presence of new training examples. One naive approach would be to completely redo the entire data analysis process from scratch on the larger data set. The current fastest method in such a case utilizes the L-BFGS quasi-Newton minimization algorithm with AllReduce along a distributed cluster, \cite{Langford:2012}. The other extreme is to apply the method of online learning, which considers one data point at a time and updates the parameters $\theta$ according to some form of gradient descent, see \cite{Langford:2009}, \cite{Duchi:2010}. However, on the tera-scale, neither approach is as appealing or as fast as we can achieve with our method. We describe in a bit more detail these recent approaches to solving (\ref{eq:costfunction}) in Section \ref{sec:prev}. For completeness, we also refer the reader to recent work relating to large-scale optimization contained in \cite{Schraudolph:2007} and \cite{Bottou:2010}. 

Our approach in simple terms lies somewhere between pure L-BFGS minimization (widely accepted as the fastest brute force optimization algorithm whenever the function is convex and smooth) and online learning. While L-BFGS offers accuracy and robustness with a relatively small number of iterations, it fails to take direct advantage of situations where the new data is not very different from that acquired previously or situations where the new data is extremely different than the old data. Certainly, one can initiate a new optimization job on the larger data set with the parameter $\theta$ initialized to the previous result. But we are left with the problem of optimizing over increasingly larger training sets at one time. Similarly, online learning methods only consider one data point at a time and cannot reasonably change the parameter $\theta$ by too much at any given step without risk of severely increasing the regret. It also cannot typically reach as small of an error count as that of a global gradient descent approach. On the other hand, we will show that it is possible to combine the advantages of both methods: in particular the small number of iterations and speed of L-BFGS when applied to reasonably sized batches, and the ability of online learning to ``forget" previous data when the new data has changed significantly.

The outline of the paper is as follows. In Section \ref{sec:background} we describe the general problem of interest. In Section \ref{sec:prev} we briefly mention current widely used methods of solving (\ref{eq:costfunction}). In Section \ref{sec:approach} we outline the statistically adaptive learning algorithm. Finally, in Section \ref{sec:experiments} we benchmark the performance of our two related methods (Context Relevant FAST L-BFGS and SA L-BFGS, respectively) against Vowpal Wabbit - one of the fastest currently available routines which incorporates the work of \cite{Langford:2012}. We also include the associated Area Under Curve (AUC) rating, which roughly speaking, is a number in $[0,1]$, where a value of $1$ indicates perfect prediction by the model.

\section{Background and Problem Setup}
\label{sec:background}
In this paper the underlying problem is as follows. Suppose we have a sequence of time-indexed data sets $\{X_{t}, Y_{t}\}$ where $X_{t} = \{x^{(i)}_{t}\}_{i=1}^{m_{t}}$, $Y_{t}= \{y_{t}^{(i)}\}_{i=1}^{m_{t}}$, $t = 0,1,\ldots, t_{f}$ is the time index, and $m_{t} \in \mathbb{N}$ is the batch size (typically independent of $t$). Such data is given sequentially as it is acquired (e.g. $t$ could represent days), so that at $t = 0$ one only has possession of $\{X_{0},Y_{0}\}$. Alternatively, if we are given a large data set all at once, we could divide it into batches indexed sequentially by $t$. In general, we use the notation $x_{t}$ with subscript $t$ to denote a time-dependent vector at time $t$, and we write $x_{t,j}$ to denote the $j$th component of $x_{t}$. For each $t = 0,\ldots, t_{f}$ we define
\begin{align}
\label{eq:lossfunction}
f_{t}(\theta) & = \sum_{i=1}^{m_{t}}l(\theta^{T}x_{t}^{(i)}; y_{t}^{(i)}) \notag\\
\phi_{t}(\theta) & = f_{t}(\theta) + \lambda S(\theta),
\end{align}
where as before, $l$ is a given smooth convex loss function and $S$ is a regularizer. Also let $\theta_{t}$ be the parameter vector obtained at time $t$, which in practice will approximately minimize $\lambda S(\theta) + \sum_{s=0}^{t}f_{s}(\theta)$. We define the regret with respect to a fixed (optimal) parameter $\theta^{*}$ (in practice we don't know the true value of $\theta^{*}$) by
\begin{align}\label{eq:regret}
R_{\phi}(t) & = r_{\phi}(\theta^{*},t) := \sum_{s=0}^{t}[\phi_{s}(\theta_{s}) - \phi_{s}(\theta^{*})] \\
& = \sum_{s=0}^{t}[f_{s}(\theta_{s}) + \lambda S(\theta_{s}) - f_{s}(\theta^{*}) - \lambda S(\theta^{*})] \notag.
\end{align}
An effective algorithm is then one in which the sequence $\{\theta_{t}\}_{t=0}^{t_{f}}$ suffers sub-linear regret, i.e., $R_{\phi}(t) = o(t)$.

As mentioned earlier, there has been much work done regarding how to solve (\ref{eq:costfunction}) with a variety of methods. Before proceeding with a basic overview of the two most popular approaches to large-scale machine learning in Section \ref{sec:prev}, it is important to understand the underlying assumptions and implications of the previous body of work.  In particular, we mention the work of L\'eon Bottou in \cite{Bottou:2010} regarding large-scale optimization with stochastic gradient descent, and \cite{Schraudolph:2007} regarding stochastic online L-BFGS (oL-BFGS) optimization. Such works and others have demonstrated that with a lot of randomly shuffled data, a variety of methods (oL-BFGS, 2nd order stochastic gradient descent and averaged stochastic gradient descent) can work in fewer iterations than L-BFGS because:
\begin{enumerate}
\item[(a)] Small data learning problems are fundamentally different from large data learning problems;
\item[(b)] The cost functions as framed in the literature have well suited curvature near the global minimum.
\end{enumerate}

We remark that the key problem for all quasi-Newton based optimization methods (including L-BFGS) has been that noise associated with the approximation process -- with specific properties dependent on each learning problem --  causes adverse conditions which can make L-BFGS (and its variants) fail. However, the problem of noise leading to non-positive curvature near the minimum can be averted if the data is appropriately shaped (i.e. feature selection plus proper data transformations). For now though, we ignore the issue and assume we already have a methodology for ``feature shaping" that assures under operational conditions that the curvature of the resulting learning problem is well-suited to the algorithm that we describe.

\section{Previous Work}
\label{sec:prev}

\subsection{Online Updates}
In online learning, the problem of storage is completely averted as each data point is discarded once it is read. We remark that one can essentially view this approach as a special case of the statistically adaptive method described in Section \ref{sec:approach} with a batch size of $1$. Such algorithms iteratively make a prediction $\theta_{t} \in \rn{l}$ and then receive a convex loss function $\phi_{t}$ as in (\ref{eq:lossfunction}). Typically, $\phi_{t}(\theta) = l(\theta^{T}x_{t}; y_{t}) + \lambda R(\theta)$, where $(x_{t},y_{t})$ is the data point read at time $t$. We then make an update to obtain $\theta_{t+1}$ using a rule that is typically based on the gradient of $l(\theta^{T}x_{t},y_{t})$ in $\theta$. Indeed, the simplest approach (with no regularization term) would be the update rule $\theta_{t+1} = \theta_{t} - \eta_{t}\nabla_{\theta}l(\theta_{t}^{T}x_{t},y_{t})$.

However, there currently exist more sophisticated update schemes which can achieve better regret bounds for (\ref{eq:regret}). In particular, the work of Duchi, Hazan, and Singer is a type of subgradient method with adaptive proximal functions. It is proven that their \textsf{ADAGRAD} algorithm can achieve theoretical regret bounds of the form
\begin{align}
\label{eq:duchiregretbound}
R_{\phi}(t) & = O(\|\theta^{*}\|_{2} \mathrm{tr}(G_{t}^{1/2}) ) \quad \textrm{ and } \\
\qquad R_{\phi}(t) & = O\left( \max_{s \leq t} \|\theta_{s} - \theta^{*}\|_{2} \mathrm{tr}(G_{t}^{1/2}) \right),\notag
\end{align}
where in general, $G_{t} = \sum_{s=0}^{t} g_{s}g_{s}^{T}$ is an outer product matrix generated by the sequence of gradients $g_{s} = \nabla_{\theta}f_{s}(\theta_{s})$ \cite{Duchi:2010}.  We remark that since the loss function gradients converge to zero under \textit{ideal} conditions, the estimate (\ref{eq:duchiregretbound}) is indeed sublinear, because the decay of the gradients, however slow, counters the linear growth in the size of $G_{t}^{1/2}$.

\subsection{Vowpal Wabbit with Gradient Descent}
Vowpal Wabbit is a freely available software package which implements the method described briefly in \cite{Langford:2012}. In particular, it combines online learning and brute force gradient-descent optimization in a slightly different way. First, one does a single pass over the whole data set using online learning to obtain a rough choice of parameter $\theta$. Then, L-BFGS optimization of the cost function is initiated with the data split across a cluster. The cost function and its gradient are computed locally and AllReduce is utilized to collect the global function values and gradients in order to update $\theta$. The main improvement of this algorithm over previous methods is the use of AllReduce with the Hadoop file structure, which significantly cuts down on communication time as is the case with MapReduce. Moreover,  the baseline online learning step is done with a learning rate chosen in an optimal manner as discussed in \cite{Langford:2011}.

\section{Our Approach}
\label{sec:approach}
\subsection{Least Squares Digression}
Before we describe the statistically adaptive approach for minimizing a generic cost function, consider the following simpler scenario in the context of least squares regression. Given data $\{X,Y\}$ with $X \in \rn{m \times l}$ and $Y \in \rn{m}$, we want to choose $\theta$ that solves $\min_{\theta \in \rn{l}}\|X\theta - Y \|_{2}^{2}$. Assuming invertibility of $X^{T}X$, it is well known that the solution is given by 
\begin{equation}\label{eq:leastsquares}
\theta = (X^{T}X)^{-1}X^{T}Y.
\end{equation}
Now, suppose that we have time indexed data $\{X_{s},Y_{s}\}_{s=0}^{T}$ with $X_{s} \in \rn{m_{s} \times l}$ and $Y_{s} \in \rn{m_{s}}$. In order to update $\theta_{t}$ given $\{X_{t+1}, Y_{t+1}\}$, first we must check how well $\theta_{t}$ fits the newly augmented data set. We do this by evaluating 
\begin{equation}
\label{eq:lserror}
\sum_{s=0}^{t+1}\|X_{s}\theta - Y_{s}\|_{2}^{2}
\end{equation}
with $\theta = \theta_{t}$. Depending on the result, we choose a parameter $\lambda \in [0,1]$ that determines how much weight to give the previous data when computing $\theta_{t+1}$. That is, $\lambda$ represents how much we would like to ``forget" the previous data (or emphasize the new data), with a value of $\lambda = 1$ indicating that all previous data has been thrown out. Similarly, the case $\lambda = \frac{1}{2}$ corresponds to the case when past and present are weighed equally, and the case $\lambda = 1$ corresponds to the case when $\theta_{t}$ fits the new data perfectly (i.e. (\ref{eq:lserror}) is equal to zero).

Let $X_{[0,t]}$ be the $\sum_{s=0}^{t}m_{s} \times l$ matrix $[X_{0}^{T},X_{1}^{T}, \ldots, X_{t}^{T}]^{T}$ obtained by concatenation, and similarly define the length-$\sum_{s=0}^{t}m_{s}$ vector $Y_{[0,t]} := [Y_{0}^{T}, Y_{1}^{T}, \ldots, Y_{t}^{T}]^{T}$. Then (\ref{eq:lserror}) is equivalent to $\|X_{[0,t+1]}\theta - Y_{[0,t+1]}\|_{2}^{2}$. Now, when using a particular second order Newton method for minimizing a smooth convex function, the computation of the inverse Hessian matrix is analogous to computing $(X_{[0,t]}^{T}X_{[0,t]})^{-1}$ above. As $t$ grows large, the cumulative normal matrix $X_{[0,t]}^{T}X_{[0,t]}$ becomes increasingly costly to compute from scratch, as does its inverse. Fortunately, we observe that
\begin{equation}
\label{eq:lsexactupdate}
X_{[0,t+1]}^{T}X_{[0,t+1]} = X_{[0,t]}^{T}X_{[0,t]} + X_{t+1}^{T}X_{t+1}.
\end{equation}
However, if we want to incorporate the flexibility to weigh current data differently relative to previous data, we need to abandon the exact computation of (\ref{eq:lsexactupdate}). Instead, letting $A_{t}$ denote the approximate analogue of $X_{[0,t]}^{T}X_{[0,t]}$, we introduce the update
\begin{equation}
\label{eq:lqupdate}
A_{t+1} \gets \frac{2}{1+\mu^{2}}\left(\mu^{2}A_{t} + X_{t+1}^{T}X_{t+1}\right)
\end{equation}
where $\mu$ satisfies $\lambda = \frac{\mu^{2}}{1+\mu^{2}}$.

The actual update of $\theta_{t}$ is performed as follows. Define 
\begin{align*}
\widetilde{Y}_{[0,t]} & := X_{[0,t]}^{T}Y_{[0,t]} = [X_{0}^{T}, \ldots, X_{t}^{T}] \left[ \begin{array}{c} Y_{0} \\ Y_{1} \\ \vdots \\ Y_{t} \end{array}\right]\\
& = \sum_{s=0}^{t}X_{s}^{T}Y_{s}.
\end{align*}
Up to time $t$, the standard solution to the least squares problem on the data $\{X_{[0,t]}, Y_{[0,t]}\}$ is then
\begin{equation}
\theta = (X_{[0,t]}^{T}X_{[0,t]})^{-1}\widetilde{Y}_{[0,t]}. \label{eq:lsnoforgetupdate}
\end{equation}
Now let $B_{t}$ be an approximation to $\widetilde{Y}_{[0,t]}$. We define $B_{t+1}$ by the update
\begin{equation*}
B_{t+1} = \frac{2}{1+\mu^{2}}\left( \mu^{2}B_{t} + X_{t+1}^{T}Y_{t+1}\right).
\end{equation*}
Finally, we set
\begin{equation}\label{eq:lsadaptiveparamupdate}
\theta_{t+1} := A_{t+1}^{-1}B_{t+1}.
\end{equation}
It is easily verified that (\ref{eq:lsadaptiveparamupdate}) coincides with the standard update (\ref{eq:lsnoforgetupdate}) when $\mu = 1$.

\subsection{Statistically Adaptive Learning}
Returning to our original problem, we start with the parameter $\theta_{0}$ obtained from some initial pass through of $\{X_{0},Y_{0}\}$, typically using a particular gradient descent algorithm. In what follows, we will need to define an easily evaluated error function to be applied at each iteration, mildly related to the cumulative regret (\ref{eq:regret}):
\begin{equation}
I(t, \theta) := \frac{\sum_{s=0}^{t}\sum_{i=1}^{m_{s}}|p_{\theta}(x_{t}^{(i)}) - y_{t}^{(i)}|}{\sum_{s=0}^{t} |X_{s}|} \label{eq:mismatch}
\end{equation}
We remark that $I(t, \theta)$ represents the relative number of incorrect predictions associated with the parameter $\theta$ over all data points from time $s=0$ to $s=t$. Moreover, because $p_{\theta}$ is a linear function of $x$, $I$ is very fast to evaluate (essentially $O(m)$ where $m = \sum_{s=0}^{t}m_{s}$).

Given $\theta_{t}$, we compute $I(t+1, \theta_{t})$. There are two extremal possibilities:
\begin{enumerate}
\item[1.] $I(t+1, \theta_{t})$ is \textit{significantly} larger than $I(t,\theta_{t})$. More precisely, we mean that $I(t+1,\theta_{t}) - I(t,\theta_{t}) > \sigma(t)$, where $\sigma(t)$ is the standard deviation of $\{I(s,\theta_{s})\}_{s=0}^{t}$. In this case the data has significantly changed, and so $\theta$ must be modified.
\item[2.] Otherwise, there is no need to change $\theta$ and we set $\theta_{t+1} = \theta_{t}$.
\end{enumerate}
\begin{figure}
\centering
\includegraphics[width=.5\textwidth]{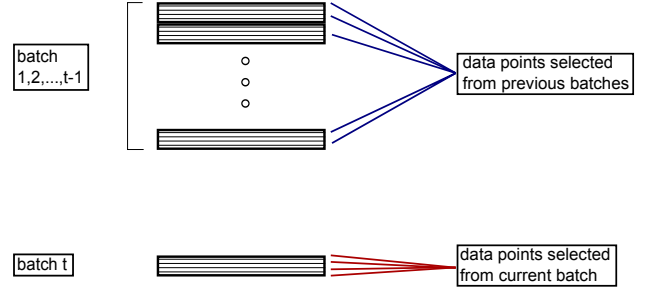}
\caption{Subsampling of the partitioned data stream at time $t$ and times $0,1,\ldots, t-1$, respectively.}\label{fig:updatediagram}
\end{figure}
In the former case, we use the magnitude of $I(t+1,\theta_{t}) - I(t,\theta_{t})$ to determine a subsample of the old and new data with $M_{\mathrm{old}}$ and $M_{\mathrm{new}}$ points chosen, respectively (see Figure \ref{fig:updatediagram}). Roughly speaking, the larger the difference the more weight will be given to the most recent data points. The sampling of previous data points serves to anchor the model so that the parameters do not over fit to the new batch at the expense of significantly increasing the global regret. This is a generalization of online learning methods where only the most recent single data point is used to update $\theta$. From the subsample chosen, we then apply a gradient descent optimization routine where the initialization of the associated starting parameters is generated from those stored from the previous iteration. In the case of L-BFGS, the rank $1$ matrices used to approximate the inverse Hessian stored from the previous iteration are used to initialize the new descent routine. We summarize the process in Algorithm \ref{alg:SALBFGS}.
\begin{algorithm}
\caption{Statistically Adaptive Learning Method (SA L-BFGS)}
\label{alg:SALBFGS}
\begin{algorithmic}[]
\Require Error checking function $I(t,\theta)$
\State Given data $\{X_{s},Y_{s}\}_{s=0}^{t_{f}}$
\State Run gradient descent optimization on $\{X_{0},Y_{0}\}$ to compute $\theta_{0}$
\For{$t = 1$ to $t_{f}$}
\If{$I(t+1,\theta_{t}) - I(t,\theta_{t}) > \sigma(t)$}
\State Choose $M_{\mathrm{old}}$ and $M_{\mathrm{new}}$
\State Subsample data
\State Run L-BFGS with initial parameter $\theta_{t}$ to obtain $\theta_{t+1}$
\Else{ $\theta_{t+1} \gets \theta_{t}$ }
\EndIf
\EndFor
\end{algorithmic}
\end{algorithm}

As a typical example, at some time $t$ we might have $M_{\mathrm{old}} = 1000$, $M_{\mathrm{new}} = 100,000$, and $\sum_{t=0}^{t_{f}}m_{t} = 1\cdot 10^{9}$. This would be indicative of a batch $\{X_{t+1}, Y_{t+1}\}$ with significantly higher error using the current parameter $\theta_{t}$ than for previously analyzed batches.

We remark that when learning on each new batch of data, there are two main aspects that can be parallelized. First, the batch itself can be partitioned and distributed among nodes in a cluster via AllReduce to significantly speed up the evaluation of the cost function and its gradient as is done in \cite{Langford:2012}. Furthermore, one can run multiple independent optimization routines in parallel where the distribution used to subsample from $X_{t+1}$ and $\cup_{s=0}^{t}X_{s}$ is varied. The resulting parameters $\theta$ obtained from each separate instance can then be statistically compared so as to make sure that the model is not overly sensitive to the choice of sampling distribution. Otherwise, having $\theta$ be too highly dependent on the choice of subsample would invalidate using a stochastic gradient descent-based approach. A bi-product of this ability to simultaneously experiment with different samplings is that it provides a quick means to check the consistency of the data.

Finally, we remark that the SA L-BFGS method can be reasonably adapted to account for changes in the selected features as new data is acquired. Indeed, it is very appealing within the industry to be able to experiment with different choices of features in order to find those that matter most, while still being able to use the previously computed parameters $\theta_{t}$ to speed up the optimization on the new data. Of course, it is possible to directly apply an online learning approach in this situation, since previous data points have already been discarded. But typical gradient descent algorithms do not a priori have the flexibility to be directly applied in such cases and they typically perform worse than batch methods such as L-BFGS\cite{Langford:2012}.

\section{Experiments}
\label{sec:experiments}

\subsection{Description of Dataset and Features}
We consider data used to predict the click-through-rate (pCTR) of online ads. An accurate model is necessary in the search advertising market in order to appropriately rank ads and price clicks. The data contains $11$ variables and $1$ output, corresponding to the number of times a given ad was clicked by the user among the number of times it was displayed. In order to reduce the data size, instances with the same user id, ad id, query, and setting are combined, so that the output may take on any positive integer value. For each instance (training example), the input variables serve to classify various properties of the ad displayed, in addition to the specific search query entered. This data was acquired from sessions of the Tencent proprietary search engine and was posted publicly on \url{www.kddcup.2012.org} \cite{kddcup:2012}.

For these experiments we build a basic model that learns from the identifiers provided in the training set.  These include unique identifiers for each query, ad, keyword, advertiser, title, description, display url, user, ad position, and ad depth (further details available in the KDD documentation).  We compute a position and depth normalized click through rate for each identifier, as well as combinations (conjunctions) of these identifiers.  Then at training and testing time we annotate each example with these normalized click through rates.  Additionally, before running the optimization, it is necessary to build appropriate feature vectors (i.e. shape the data). We will not go into detail regarding how this is done, except to mention that the number of features generated is on the order of $1000$.

\subsection{Model 1 Results}
For our first set of experiments, we compare the performance of Vowpal Wabbit (VW) using its L-BFGS implementation and the Context Relevant Flexible Analytics and Statistics Technology\texttrademark{} L-BFGS implementation running on $10$ Amazon m1.xlarge instances. The time measured (in seconds) is only the time required to train the models. The features are generated and cached for each implementation prior to training. 

Performance was measured using the Area Under Curve (AUC) metric because this was the methodology used in  \cite{kddcup:2012}. In short, the AUC is equal to the probability that a classifier will rank a randomly chosen positive instance higher than a randomly chosen negative one. More precisely, it is computed via Algorithm 3 in \cite{fawcett:2004}. We compute our AUC results over a portion of the public section of the test set that has about 2 million examples.

Model 1 includes only the basic id features, with no conjunction features, and achieves an AUC of 0.748 as shown in Table \ref{model1}. A simple baseline performance, which can be generated by predicting the mean ctr for each ad id would perform at approximately an AUC of 0.71. The winner of  \cite{kddcup:2012} performed at an AUC of 0.80. However, the winning model was substantially more complicated and used many additional features that were excluded from this simple demonstration. In future work, we will explore more sophisticated feature sets.

The Context Relevant and VW models both achieve the same AUC on the test set, which validates that the basic gradient descent and L-BFGS implementations are functionally equivalent.  The Context Relevant model completes learning between four and five times more quickly.  Our implementation is heavily optimized to reduce computation time as well as memory footprint.  In addition, our implementation utilizes an underlying MapReduce implementation that provides robustness to job and node failures\footnote{Context Relevant had to re-write the AllReduce network implementation to add error checking so that the AllReduce system was robust to errors that were encountered during normal execution of experiments on Amazon's EC2 systems. Without these changes, we could not keep AllReduce from hanging during the experiments. There is no graceful recovery from the loss of a single node.}.

\begin{table}[ tp ]%
\caption{Model 1 Results For Different Learning Mechanisms (VW = Vowpal Wabbit; CR = Context Relevant  FAST L-BFGS}
\label{model1}\centering %
\rowcolors{3}{gray!35}{gray!10}
\begin{tabular}{ccccc}

        & VW     & CR     \\
        & L-BFGS & L-BFGS \\
\hline
seconds &  490 &  114\\
AUC     & .748 & .748\\
\hline
\end{tabular}
\end{table}

\subsection{Model 2 Results}
Context Relevant's implementation of SA L-BFGS is designed to accelerate and simplify learning iterative changes to models. Using information gleaned from the initial L-BFGS pass, SA L-BFGS develops a sampling strategy to minimize sampling induced noise when learning new models that are derived from previous models. The larger the divergence in the models, the less speed-up is likely. For this set of experiments, we add a conjunction feature that captures the interaction between a query id and an ad id, which has frequently been an important feature in well known advertising systems.  We then compare the speed and accuracy of Vowpal Wabbit (VW) using its L-BFGS implementation; the Context Relevant Flexible Analytics and Statistics Technology\texttrademark{} L-BFGS implementation, and the Context Relevant Flexible Analytics and Statistics Technology\texttrademark{} SA L-BFGS implementation (SA) running on $10$ Amazon X1.Large instances. Here the baseline L-BFGS models are trained with the standard practice for adding a new feature, the models are retrained on the entire dataset.  The time measured (in seconds) is only the time required to train the models. The features are generated and cached for each implementation prior to training. 

Again, performance was measured using the Area Under Curve (AUC) metric. Table \ref{model2} lists the results for each algorithm, and shows that AUC improves in comparison to Model 1 when the new feature is added. As with Model 1, the VW and CR models achieve similar AUC and the basic L-BFGS CR learning time is significantly faster.  Furthermore, we show that SA L-BFGS also achieves similar AUC, but in less than one tenth of the time (which likewise implies one tenth of the compute cost required). This speed up can enable a large increase in the number of experiments, without requiring additional compute or time.  It is important to note that the speed of L-BFGS and SA L-BFGS is essentially tied to the number of features and the number of examples for each iteration. The primary performance gains that can be found are: (a) reducing the number of iterations; (b) reducing the number of examples; or (c) reducing the number of features with non-zero weights. SA L-BFGS adopts the former two strategies. A reduction in the number of features with non-zero weights can be forced through aggressive regularization, but at the expense of specificity.

\begin{table}[ tp ]%
\caption{Model 2 Results For Different Learning Mechanisms (VW = Vowpal Wabbit; CR = Context Relevant  FAST L-BFGS; SA = Context Relevant FAST SA L-BFGS)}
\label{model2}\centering %
\rowcolors{3}{gray!35}{gray!10}
\begin{tabular}{ccccc}

        & VW     & CR     & CR\\
        & L-BFGS & L-BFGS & SA L-BFGS\\
\hline
seconds &  515 & 115 & 9 \\
AUC     & .750 & .752 & .751 \\
\hline
\end{tabular}
\end{table}

\section{Conclusions}
\label{sec:conclusions}
We have presented a new tera-scale machine learning system that enables rapid model experimentation. The system uses a new version of L-BFGS to combine the robustness and accuracy of second order gradient descent optimization methods with the memory advantages of online learning.  This provides a model building environment that significantly lowers the time and compute cost of asking new questions.  The ability to quickly ask and answer experimental questions vastly expands to set of questions that can be asked, and therefore the space of models that can be explored to discover the optimal solution. 

SA L-BFGS is also well suited to environments where the underlying distribution of the data provided to a learning algorithm is shifting. Whether the shift is caused by changes in user behavior, changes in market pricing, or changes in term usage, SA L-BFGS can be empirically tuned to dynamically adjust to the changing conditions.  One can utilize the parallelized approach in \cite{Langford:2012} on each batch of data in the time variable, with the additional freedom to choose the batch size. Furthermore, the statistical aspects of the algorithm provide a useful way to check the consistency of the data in real time.  However, like all L-BFGS implementations that rely on small, reduced, or sampled data sets, increased sampling noise from the L-BFGS estimation process affects the quality of the resulting learning algorithm. Users of this new algorithm must take care to provide smooth convex loss functions for optimization. The Context Relevant Flexible Analytics and Statistics Technology\texttrademark{} is designed to provide such functions for optimization.

\section{About Context Relevant and the Authors}
Context Relevant was founded in March 2012 by Stephen Purpura and Christian Metcalfe. The company was initially funded by friends, family, Seattle angel investors, and Madrona Venture Group. 

Stephen Purpura -- CEO \& Co-Founder -- Stephen works as the CEO and CTO of the company. He has more than 20 years of experience, is listed as an inventor of five issued United States patents and served as program manager of Microsoft Windows and the Chief Security Officer of MSFDC, one of the first Internet bill payment systems. Stephen is recognized as a leading expert in the fields of machine learning, political micro-targeting and predictive analytics.

Stephen received a bachelor's degree from the University of Washington, a master's degree from Harvard, where he was part of the Program on Networked Governance at the John F. Kennedy School of Government, and he is set to be granted a PhD in information science from Cornell.

Jim Walsh -- VP Engineering -- Jim brings 23 years of experience in technology innovation and engineering management to Context Relevant. He founded, built and led the Cosmos team -- Microsoft's massive scale distributed data storage and analysis environment, underpinning virtually all Microsoft products including Bing -- and founded the Bing multimedia search team. In his final position at Microsoft, Jim served as the principal architect for Microsoft advertising platforms.

Prior to joining Microsoft, Jim created two software startups, wrote the second-ever Windows application and created custom computer animation software for the television broadcast industry. He is recognized as one of the world's leading network performance experts and has been granted twenty two technology patents that range from performance to user interface design. Jim earned a bachelor of science degree in computing science from the University of Alberta.

Dustin Rigg Hillard -- Director of Engineering and Data Scientist -- Dustin is a recognized data science and machine learning expert who has published more than 30 papers in these areas. Previously at Microsoft and Yahoo!, he spent the last decade building systems that significantly improve large-scale processing and machine-learning for advertising, natural language and speech.

Prior to joining Context Relevant, Dustin Hillard worked for Microsoft, where he worked to improve speech understanding for mobile applications and XBox Kinect. Before that he was at Yahoo! for three years, where he focused on improving ad relevance.  His research in graduate school focused on automatic speech recognition and statistical translation.  Dustin incorporates approaches from these and other fields to learn from massive amounts of data with supervised, semi-supervised, and unsupervised machine learning approaches.

Dustin holds a bachelor's and master's degree as well as a PhD in Electrical Engineering from the University of Washington.

Scott Golder -- Data Scientist \& Staff Sociologist -- Scott mines social networking data to investigate broad questions such as when people are happiest (mornings and weekends) and how Twitter users form new social ties. His work has been published in the journal Science as well as top computer science conferences by the ACM and IEEE, and has been covered in media outlets such as MSNBC, The New York Times, The Washington Post and National Public Radio. He has also been profiled by LiveScience's ``ScienceLives''.

He has worked as a research scientist in the Social Computing Lab at HP Labs and has interned at Google, IBM and Microsoft. Scott holds a master's degree from MIT, where he worked with the Media Laboratory's Sociable Media Group, and graduated from Harvard University, where he studied Linguistics and Computer Science. Scott is currently on leave from the PhD program in Sociology at Cornell University. 

Mark Hubenthal -- Member of the Technical Staff -- Mark recently received his PhD in Mathematics from the University of Washington. He works on inverse problems applicable to medical imaging and geophysics. 

Scott Smith - Principal Engineer and Architect - Scott has experience with distributed computing, compiler design, and performance optimization.  At Akamai, he helped design and implement the load balancing and failover logic for the first CDN.  At Clustrix, he built the SQL optimizer and compiler for a distributed RDBMS.  He holds a masters degree in computer science from MIT.

\bibliographystyle{acl}
\bibliography{fastreduce}
\end{document}